\documentclass{article} 
\usepackage{onecol}

\usepackage{times}
\usepackage{microtype}
\usepackage{epsfig}
\usepackage{graphicx}
\usepackage{amsmath}
\usepackage{amssymb}
\usepackage{algorithm}
\usepackage{algorithmic}

\usepackage{color}
\usepackage{ccicons}    
\usepackage{txfonts}
\usepackage{textcomp}

\usepackage{url}
\usepackage{multirow}
\usepackage{booktabs}
\usepackage{mathtools}
\usepackage{tabularx}
\usepackage{varwidth}
\usepackage{stmaryrd}
\usepackage{balance}

\usepackage[toc,page]{appendix}
\usepackage{makecell}
\usepackage[symbol]{footmisc}

\newcommand{\vnudge}{\vspace{-.13in}}
\renewcommand{\thefootnote}{\fnsymbol{footnote}}

\newcommand{\R}{\mathbb{R}}

\makeatletter
\DeclareRobustCommand\onedot{\futurelet\@let@token\@onedot}
\def\@onedot{\ifx\@let@token.\else.\null\fi\xspace}

\makeatother

\usepackage[pagebackref=true,breaklinks=true,letterpaper=true,colorlinks,bookmarks=false]{hyperref}
\begin{document}
\newcommand\blfootnote[1]{%
  \begingroup
  \renewcommand\thefootnote{}\footnote{#1}%
  \addtocounter{footnote}{-1}%
  \endgroup
}
\title{Rank3DGAN: Semantic mesh generation using relative attributes}

\author{Yassir~Saquil$^*$, Qun-Ce~Xu$^*$, Yong-Liang~Yang, and Peter~Hall\\
University of Bath \\
}
\date{}

\maketitle
\begin{abstract}\noindent
In this paper, we investigate a novel problem of using generative adversarial networks in the task of 3D shape generation according to semantic attributes. Recent works map 3D shapes into 2D parameter domain, which enables training Generative Adversarial Networks (GANs) for 3D shape generation task. We extend these architectures to the conditional setting, where we generate 3D shapes with respect to subjective attributes defined by the user. Given pairwise comparisons of 3D shapes, our model performs two tasks: it learns a generative model with a controlled latent space, and a ranking function for the 3D shapes based on their multi-chart representation in 2D. The capability of the model is demonstrated with experiments on HumanShape, Basel Face Model and reconstructed 3D CUB datasets. We also present various applications that benefit from our model, such as multi-attribute exploration, mesh editing, and mesh attribute transfer.
\blfootnote{$^*$ Equal contribution}
\end{abstract}

\section{Introduction}

Building a generative model for 3D shapes has a wide variety of applications in computer vision and graphics, in both research and industrial fields. For instance, we can design a system for a computer animator so that he can manipulate 3D shapes not only based on low-level attributes such as geometry, but also high-level semantic attributes such as expressions for 3D faces or girth for 3D human bodies.

However, learning a generative model for fine-grained 3D shapes with high quality remains a challenging problem. The current state-of-the-art methods rely on representing the 3D shape as a tensor data itself which begets many limitations. For volumetric grid ({\em i.e.}~voxel) representation \cite{V1,V2,V3}, surface details such as smoothness and continuity are lost due to limited resolution of voxel tensors. In contrast, point cloud \cite{PC1,PC2,PC3} representation is simple and straightforward, but it is not a regular structure to fit into neural networks and it does not model connectivity between points, making 3D surface reconstruction a non-trival task. These drawbacks hinder the extension of these methods to more sophisticated tasks such as, semantic manipulation of quality mesh models.

Recently, Hamu {\em et al.} \cite{Multi-Chart} proposed to represent 3D mesh surfaces using 2D multi-chart structure based on orbifold tutte embedding \cite{Tutte}. The basic idea is to cut the mesh (or part of it) through landmark points on the surface, then map it to a chart structure in 2D, and finally sample a grid in 2D. This results in a tensor of image-like data, which is both smooth and bijective. Moreover, the convolution operation is well-defined and can be mapped back to the original surface. Hence a Generative Adversarial Networks \cite{GAN} can be trained based on such multi-chart structure to generate multi-channel images, which implicitly represent a 3D mesh surface that can be easily reconstructed.

Inspired by the semantic image manipulation using RankCGAN \cite{RankCGAN}, our aim is to extend this work to 3D mesh surfaces using the multi-chart structure \cite{Multi-Chart}, which enables us to provide a system that can generate 3D meshes while manipulating some semantic attributes defined by the user. The key characteristic of this approach is the addition of a ranking module that can order 3D shapes via their multi-chart representation.

We evaluated our method on three types of datasets of human bodies (HumanShape \cite{HumanShape}), birds (reconstructed 3D shapes of CUB imagery data \cite{Bird}), and human faces (Basel Face Model \cite{Basel09,Basel17}). Both quantitative and qualitative results show that: our model can disentangle semantic mesh attributes while ensuring a coherent variation of the meshes 
and also benefit several important applications in mesh generation, editing, and transfer.

Our contributions are two-fold: 1) a novel conditional generative model that can generate quality 3D meshes with semantic attributes quantified by a ranking function; 2) a training scheme that only requires pairwise comparison of meshes instead of global annotation of the whole dataset.

\section{Related work}

\paragraph{3D Shape Generative Models:}
Early works on 3D shape synthesis relied on probabilistic inference methods. \cite{PM2,PM3} propose a probabilistic generative model of component-based 3D shapes, while \cite{PM1} reconstructed 3D geometry from 2D line drawings using maximum-a-posteriori (MAP) estimate. These models are generally adequate for a certain class of 3D shapes.

Recent works shifted towards using deep generative models for 3D data synthesis. These works can be categorized according to the learned 3D representation, which fall into three types: voxel, point clouds and depth map based methods.
Unlike the component-based methods, the voxel-based methods do not rely on part labeling and learn voxel grids from a variety of inputs; a probabilistic latent space \cite{3DGAN,3DBN}, single or multi-view images \cite{3D_multi,3D_image,3D_image1,3D_image2}, noisy or part-missing shapes \cite{3D_completion,3D_denoise}. However, generated 3D voxels are generally low resolution and memory costly. To mitigate this issue, recent works \cite{octree1,octree2,octree3} investigated the possibility of using the octree representation.

Another line of work focused on generating 3D point clouds \cite{PC3,PC4}. This representation avoids the resolution problem, but induces challenges to ensure the order and transformation invariance of the 3D points \cite{PC1,PC2}. Another limitation of point clouds is the lack of point connectivity, which complicates the task of reconstructing a 3D mesh.

Finally, some works considered generating depth maps \cite{Depth}, which are usually used as an intermediate step along with an estimated silhouette or normal map to generate a 3D shape \cite{Depth_3D1,Depth_3D2,Depth_3D3}.

\paragraph{Mesh Generation and Processing:} 
There has been an increasing interest to generalize deep learning operators to curved surface meshes. \cite{Surface1,Surface2} formulated the convolution operation in the non-Euclidean domain as template matching with local patches in geodesic or diffusion system. \cite{Surface3} proposed a mixture model networks on graphs and surfaces and built parametric local patches instead of fixed ones. These methods were demonstrated in the tasks of shape correspondence and description.

In another spectrum, since triangular meshes are widely used representation of 3D shapes, many works processed meshes with neural networks despite their irregular format. \cite{VAE_3D1,VAE_3D2} proposed a variational autoencoder to encode meshes using rotation-invariant representation \cite{Gao1,Gao2} for shape embedding, synthesis and deformation transfer. In addition, few works considered the 3D mesh as a graph structure. \cite{shape_completion} used a graph variational auto-encoder for deformable shapes completion. \cite{pixel2mesh} learned a geometric deformation of an ellipsoid from a single image using graph-based CNN \cite{GraphNet}. The main limitation of these works is the one-to-one correspondences requirement of the input data.

Surface parameterization for mapping 3D surfaces to a 2D domain is a well studied problem in computer graphics. \cite{Geo_CNN} learned CNN models using geometry images \cite{Geo_Img}. However, geometry images are neither seamless nor unique, making the convolution operation not translation-invariant. \cite{Toric} tackled these issues by parameterizing a 3D surface to a global seamless flat-torus, where the convolution is well defined. These methods are restricted to sphere-type shapes and are applied to shape classification, segmentation and landmark detection tasks.

For 3D mesh surface synthesis, \cite{Geo_Gen} learned to generate geometry images as a shape representation. On the other hand, AtlasNet \cite{AtlasNet} used MLPs to learn multiple parameterizations that map 2D squares with latent shape features to the 3D surface. Despite no restriction on surface topology, the generated shapes have less details and rely on an input image or point cloud, thus limit further applications. Meanwhile, based on a sparse set of surface landmarks, \cite{Multi-Chart} represented a 3D shape by a collection of conformal charts formed by flat-torus parameterization \cite{Toric}. This reduces mapping distrotion and can be used for multiple tasks that requires mesh quality, notably mesh generation using GANs \cite{GAN}. Our work is an extension of multi-chart 3D GAN \cite{Multi-Chart} in the conditional setting using relative attributes.

\paragraph{Relative Attributes:}

In early studies, binary attributes that indicate the presence or the absence of an attribute in data showed state-of-the-art performance in object recognition \cite{Recognition} and action recognition \cite{Action}. It allowed prediction of color or texture types \cite{CatAtt}, zero-shot learning \cite{RussakovskyF10} and part localization \cite{CatAtt2}. These attributes can be extracted from the Web and knowledge resources \cite{Relative}

However, a better representation of attributes is necessary if we want to quantify the emphasis of an attribute and compare it with others. For this reason, relative attributes \cite{Relative} tackled this issue by learning a global ranking function on data using constraints describing the relative emphasis of attributes ({\em e.g.}~pairwise comparisons of data). This approach is regarded as solving a learning-to-rank problem where a linear function is learned based on RankSVM \cite{RankSVM}. This problem can also be modeled by a non-linear function as in RankNet \cite{RankNet}, where the ranker is a neural network trained using gradient descent methods. Furthermore, semi-supervised learning can as well learn user-defined ranking functions as demonstrated in criteria sliders \cite{Tompkin:2017:BMVC}.

These algorithms focus on predicting attributes on existing data entries. RankCGAN \cite{RankCGAN} proposes an extension enabling to continuously synthesize new imagery data according to semantic attributes. Our work can also be treated as an extension to 3D mesh synthesis with respect to subjective criteria.

\section{Approach}

In this section, we first outline the method of adapting 2D CNN architectures to 3D mesh surfaces through mesh parameterization \cite{Toric} \cite{Multi-Chart}, then we show how to make extension in the conditional setting for semantic mesh generation using relative attributes.

\subsection{Mesh parameterization for 2D CNN adaptation}
Building a generative model for 3D meshes consists of learning a function $G: \R^d \to \mathcal{S}$ that maps a latent distribution in $\R^d$, usually a gaussian or uniform noted by $p_z$, to a complex distribution $p_g$ in $\mathcal{S}$ that approximates the real distribution $p_x$ of our training surfaces $\{S_i\}_{i \in I}$. Where $d$ is the dimension of the latent space and $\mathcal{S}$ is the surface space.

GANs \cite{GAN} are popular generative models for 2D images. But CNN architectures cannot be applied directly on data sampled from $\mathcal{S}$, which shows the necessity of transferring the learning problem from surface to image space where the convolution operation is defined. For this reason, \cite{Toric} suggest to conformally parameterize the mesh surface to a flat-torus by picking a triplet of landmark points. The flat-torus can be represented as a planar square, which is handy because we can discretize it to a $n \times n $ grid and apply standard 2D convolutions directly on the sampled grid.

However, as conformal parameterization only minimizes angle distortion, the global flat-torus structure may result in significant scale distortion, which largely affects the quality of the reconstructed 3D shape with perturbing parts ({\em e.g.}~human body). This is because 2D grid discretization is performed uniformly, which can easily lead to severe geometry loss at highly distorted region. To address this, a multi-chart structure composed by a set of charts that globally cover the mesh surface is proposed to reduce the scale distortion \cite{Multi-Chart}. Multiple landmark points need to be specified to define the multi-chart structure, where a single chart is constructed using a triplet/quadriplet of landmarks as in \cite{Tutte,Toric}.

The above parameterization based method allows to adapt 2D CNNs for generating quality 3D meshes with relative attributes. More detailed explanation can be found in the Appendix.

\subsection{Semantic Mesh Generation}

\begin{figure*}[t]
\includegraphics[width=0.7\linewidth,trim={0 0 0 0}, clip]{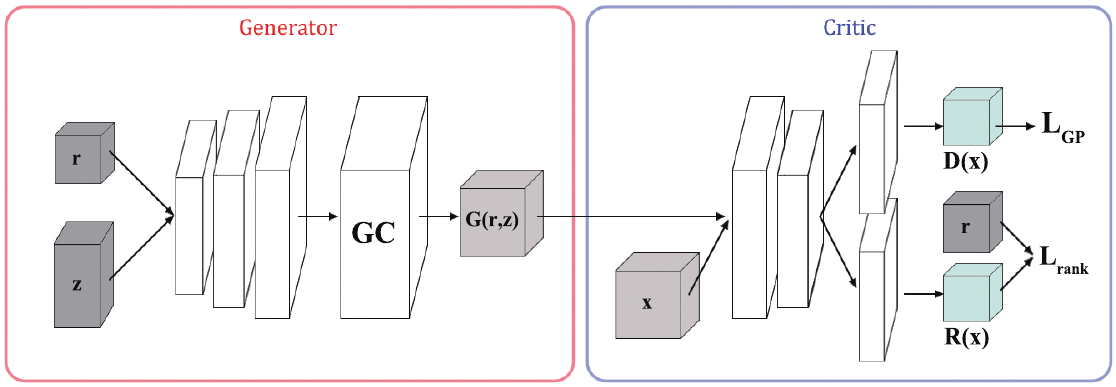}
\centering
\caption{Rank3DGAN network: The dark grey represents the latent variables, light grey represents the chart images, light blue indicates the outputs of the critic, and the dark blue indicates their loss functions. GC represents the geometric consistency layers detailed in Appendix.}
\label{fig:GAN}
\vnudge
\end{figure*}



We recall that Generative Adversarial Networks (GANs) \cite{GAN} training is defined as a minmax game between two networks; A generator $G$ that maps a latent variable $z \sim p_z$ to generated sample $G(z)$, and a discriminator $D$ that classifies an input sample as either a real or generated sample. Due to the training instability with the former loss, WGAN \cite{WGAN} proposes using Wasserstein distance defined as the minimum cost of transporting mass to transform the generated distribution into real distribution. 
The WGAN loss suggests clipping the $D$ (called a \emph{critic}) weights to enforce the Lipschitz constraint.   

WGAN-GP \cite{WGAN-GP} proposes another alternative for setting the Lipschitz constraint, which consists of a penalty on the gradient norm, controlled with a hyperparameter $\lambda = 10$, for random samples $\hat{x} \sim p_{\hat{x}}$  with $p_{\hat{x}}$ a uniform sampling along straight lines between samples from real $p_x$ and generated $p_g$ distributions. The WGAN-GP \cite{WGAN-GP} optimizes the following loss:
\begin{equation}
    \mathcal{L}_{GP} = \mathbb{E}_{\Tilde{x} \sim p_g} [D(\Tilde{x})] - \mathbb{E}_{x \sim p_x} [D(x)] + \lambda~\mathbb{E}_{\hat{x} \sim p_{\hat{x}}}[(||\nabla_{\hat{x}} D(\hat{x}) ||_2 - 1)^2]
\end{equation}

In order to generate meshes controlled by one or more subjective attributes, we need to incorporate semantic ordering constraints in WGAN-GP loss. We follow \cite{RankCGAN} in modelling the semantic constraints by a ranking function, noted $R$, to be incorporated in the training of critic and generator. Training a pairwise ranker $R$ using CNNs was proposed in \cite{RankNet,DeepRel}, which consists of learning to classify a pair of inputs, $x^{(1)}$ and $x^{(2)}$ according to their ordering, $x^{(1)} \succ x^{(2)}$ or $x^{(1)} \prec x^{(2)}$. Formally, given a dataset $\{(x_i^{(1)}, x_i^{(2)}, y)\}_{i=1}^P$ of size $P$, such that $(x_i^{(1)},x_i^{(2)})$ pair of images and $y_i \in \{0,1\}$ a binary label indicating if $x_i^{(1)} \succ x_i^{(2)}$ or not, we define the ranking loss as follows:
\begin{equation}
    \mathcal{L}_{rank}(x_i^{(1)},~x_i^{(2)},~y_i)
    = -y_i \log[\sigma (R(x^{(1)}_i)-R(x^{(2)}_i))]
    - (1-y_i) \log[1-\sigma (R(x^{(1)}_i)-R(x^{(2)}_i))]),
\end{equation}
\\
with $\sigma(x) = \frac{1}{1 + e^{-x}}$ the sigmoid function. We note that the ranker $R$ can be regarded as a function that outputs the ranking score of the input data.

With all necessary components presented, we propose an architecture, called Rank3DGAN, that can generate meshes according to subjective attributes set by the user. The underlying mechanism is the addition of semantic ordering constraint, modelled by a ranking function, to the generative model trained using multi-chart structure representing the 3D meshes data.

A fundamental difference between our work and \cite{RankCGAN} is in the application context, since we focus on manipulating 3D shapes via the multi-chart structure. We also opted for architecture changes. As illustrated in Figure \ref{fig:GAN}, instead of having a separate network for the ranker as in \cite{RankCGAN}, we introduce the ranker in the critic network $D$ as an auxiliary classifier \cite{AuxClass}, so that the critic $D$ has two outputs for adversarial and ranking losses respectively. Having the ranker as an auxiliary classifier ensures that the training is end-to-end and the ranker can benefit from the learned representation of the critic.

In the following subsections, we describe Rank3DGAN in the mesh generation task with respect to one attribute, then we generalize to multiple attributes.  Finally, we introduce an inference network that estimates the latent variables of an input multi-chart tensor for the task of mesh editing.

\subsubsection{One attribute generation}

As shown in Figure \ref{fig:GAN}, the generator $G$ outputs a multi-chart tensor $G(r,z)$ given two inputs, the noise latent vector $z \sim \mathcal{N}(0,1)$ and the attribute latent variable $r \sim \mathcal{U}(-1,1)$. The critic $D$ takes a real $I=x$ or fake $I=G(r,z)$ multi-chart tensor and outputs its critic value $D(I)$ and ranking score $R(I)$. Given $\{(x_i^{(1)}, x_i^{(2)}, y_i)\}_{i=1}^P$ and $\{x_i\}_{i=1}^N$, mesh pairwise comparisons and mesh datasets represented by their multi-chart structure, with size of $P$ and $N$ respectively, we train Rank3DGAN in mini-batch setting of size $B$ by defining the generator $(\mathcal{L}_G)$ and critic $(\mathcal{L}_D)$ loss functions as follows:

\begin{equation}
\label{eq:1}
    \mathcal{L}_{D} = \frac{1}{B} \sum_{i=1}^B [D(\Tilde{x_i}) - D(x_i) + \lambda~(||\nabla_{\hat{x_i}} D(\hat{x_i}) ||_2 - 1)^2] + \frac{2}{B} \sum_{i=1}^{B/2} \mathcal{L}_{rank}(x_i^{(1)}, x_i^{(2)}, y_i),
\end{equation}

\begin{equation}
\label{eq:2}
    \mathcal{L}_{G} = -\frac{1}{B} \sum_{i=1}^B D(\Tilde{x_i}) + \nu~\frac{2}{B} \sum_{i=1}^{B/2} \mathcal{L}_{rank}(\Tilde{x}_i^{(1)}, \Tilde{x}_i^{(2)}, [r_i^{(1)} > r_i^{(2)} ]),
\end{equation}
\\
with $\Tilde{x}_i = G(r_i, z_i),~\hat{x}_i = \alpha \Tilde{x}_i + (1-\alpha) x_i,$ for $\alpha \sim \mathcal{U}(0,1)$, $\Tilde{x}_i^{(1)} = G(r_i^{(1)},z_i^{(1)}),~\Tilde{x}_i^{(2)} = G(r_i^{(2)}, z_i^{(2)})$. $[.]$ is Iverson bracket. $\nu$ (hyperparameter) controls the ranker gradient magnitude in the generator update.

\subsubsection{Multiple attributes generation}

Our method could be extended to the case of multiple attributes where the attribute latent space is vector representing the set of controllable attributes and the critic network $D$ outputs a vector of ranking scores with respect to each attribute. Formally, let $\{(x_i^{(1)}, x_i^{(2)},\mathbf{y_i})\}_{i=1}^P$ mesh pairwise comparisons dataset with $\mathbf{y_i}$ a binary vector indicating whether $x^{(1)}_i \succ x^{(2)}_i$ or not with respect to all attributes $A$. The new ranking loss is defined as follows:
\begin{equation}
    \mathcal{L}_{m-rank}(x_i^{(1)}, x_i^{(2)},\mathbf{y_i}) = -\sum_{j=1}^{A} y_{ij} \log[\sigma (R_j(x^{(1)}_i)-R_j(x^{(2)}_i))]
    + (1-y_{ij}) \log[1-\sigma (R_j(x^{(1)}_i)-R_j(x^{(2)}_i))])
\end{equation}
\\
with $y_{ij},~R_j(x^{(1)}_i),~R_j(x^{(2)}_i)$ the j-th elements in $\mathbf{y}_i,~R(x^{(1)}_i),~R(x^{(2)}_i)$ vectors respectively. By substituting $\mathcal{L}_{rank}$ in Equations \ref{eq:1} and \ref{eq:2} with $\mathcal{L}_{m-rank}$, we obtain the generative model with multiple attributes.

\subsubsection{Encoder for mesh editing}


The last component to add to our generative model, is an inference network for tasks that require estimating the latent variables of an input mesh such as mesh editing or attributes transfer. Following the previous work \cite{IGAN} in image manipulation, we propose to train the encoder $E$ on the real multi-chart tensors dataset $\{x_i\}_{i=1}^N$ in order to estimate the latent variables that minimise the reconstruction loss defined in the mini-batch setting as $\mathcal{L}_{E} = \frac{1}{B} \sum_{i=1}^{B} \parallel G(E(x_i)) - x_i  \parallel_2^2$.

\section{Empirical results}
\label{experiments}
In this section, we describe our experimental settings, show quantitative and qualitative results, and then demonstrate applications in mesh generation, mesh editing, and mesh attribute transfer.

\subsection{Data preparation}

We relied on four datasets for our experiments: 3D human shapes from MPII Human Shape \cite{HumanShape}, 3D bird meshes reconstructed from CUB image dataset \cite{Bird}, and 3D faces from Basel Face Model 2009 \cite{Basel09} and 2017 \cite{Basel17}. We sampled a large variety of meshes from each dataset with different attributes, and formed pairs with ordered attributes. For the meshes within each dataset, we also consistently distributed a good number of landmarks with triplet/quadriplet relations to form the chart-based structure, allowing the generative model to work on 3D meshes.  Table~\ref{tab:dataset} summarizes the information of all the datasets. Details on data preparation can be found in the Appendix.
\begin{table*}[h]
\centering{
\begin{tabular}{|c||c|c|c|c|c|}
  \hline
  Dataset & \# meshes & attributes & \# pairs & \# lmks & \# triplets \\
  \hline
  \hline
  MPII Human Shape & 15,000 & weight, gender & 5,000 & 21 & 16 \\
  \hline
  CUB-200-2011 & 5,000 & lifted wings & 5,000 & 14 & 11 \\
  \hline
  Basel Face Model 09 & 5,000 & gender, height, weight, age & 5,000 & 4 & N/A \\
  \hline
  Basel Face Model 17 & 5,000 & \makecell{anger, disgust, fear, \\ happy, sad, surprise} & 5,000 & 4 & N/A \\
  \hline
\end{tabular}
}
\caption{We prepared data from four datasets for our experiments. `lmks' stands for landmarks.
}
\label{tab:dataset}
\end{table*}

\subsection{Implementation}

We implemented Rank3DGAN on top of multi-chart 3D GAN \cite{Multi-Chart} with the same hyperparameter setting. The structure of the critic $D$ is modified to adapt for the incorporated ranker as in Figure \ref{fig:GAN}, while the structure of the generator $G$ remains intact.
We modelled the encoder $E$ with the same architecture of the vanilla critic $D$.
For the multi-chart representation, we sampled $64 \times 64$ images from human and bird flattened meshes, and $128 \times 128$ for faces. For the training, we set the hyperparameters $\lambda = 10$, $\nu = 1$, and trained the networks for $300$ epochs on all datasets. Similarly to \cite{Multi-Chart}, we activated the landmark consistency layer after $50$ training epochs for human and bird datasets. This layer is not used for faces as they are represented by a single chart. 
Finally, we obtained face textures using the nearest neighbors in the real face dataset for better rendering of the results.

\subsection{Quantitative Results}

\begin{figure*}[t]
\includegraphics[width=\linewidth,trim={0 0 0 0}, clip]{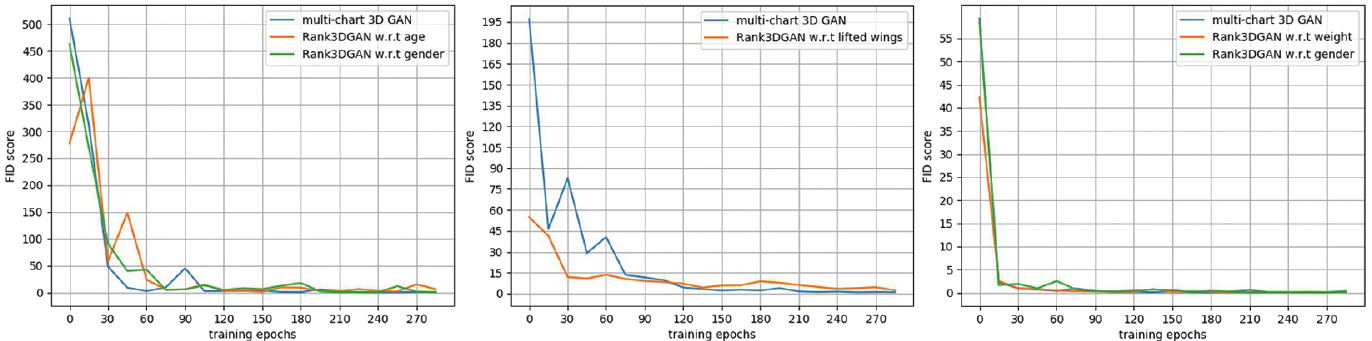}
\centering
\vnudge
\caption{Mean-FID on face, bird and human datasets. \textbf{Left:} face dataset trained on \emph{age} and \emph{gender}. \textbf{Middle:} bird dataset trained on \emph{lifted wings}. \textbf{Right:} human dataset trained on \emph{gender} and \emph{weight}. }
\label{fig:FID}
\vnudge
\end{figure*}

A common metric for evaluating the quality of generated images by GANs is Fr\'echet Inception Distance (FID) \cite{FID}. It consists of calculating the Fr\'echet Distance of two multivariate Gaussians that are estimated from two datasets of real and generated image features. To use this metric in our tasks, we modelled the feature extraction network by FCN \cite{FCN} and calculated FID with respect to each chart in the multi-chart structure, then we averaged the obtained values to get the final mean-FID result.

We trained FCN \cite{FCN} for segmentation task in face, human, bird datasets. 
Basel Face Model \cite{Basel09} provides 4 segment labels along with the face morphable model. We sampled $5000$ meshes whose coordinates and labels are represented using the multi-chart structure to create a dataset of $\{(C_i,l_i)\}_{i=1}^{5000|F|}$, with $|F|=1$ for one quadriplet, $C_i \in \R^{3 \times 128 \times 128}$ and $l_i \in [0 \mathrel{{.}\,{.}}\nobreak 3]^{128 \times 128}$ for the coordinates and label chart. For Human Shape \cite{HumanShape} and Bird \cite{Bird} datasets, we manually labelled the mesh vertices and sampled $5000$ meshes, resulting in datasets of $\{(C_i,l_i)\}_{i=1}^{5000|F|}$ with $C_i \in \R^{3 \times 64 \times 64}$, $|F|=16$, $l_i \in [0 \mathrel{{.}\,{.}}\nobreak 7]^{64 \times 64}$ for human meshes, and $|F|=11$, $l_i \in [0 \mathrel{{.}\,{.}}\nobreak 4]^{64 \times 64}$ for bird meshes.

Once FCN \cite{FCN} is trained for segmentation task, we extract features of $2048$ dimensions from conv7 layer, modified for this purpose, using input real and generated chart images of the same size to build two sets of real and generated features. To investigate the variation in the quality of generated images, we compared the calculated mean-FID of Rank3DGAN with multi-chart 3DGAN \cite{Multi-Chart} at every $15$ epochs on all the datasets (see Figure \ref{fig:FID}).
Note that the mean-FID value differences in-between becomes less significant as the models progress through the training epochs, indicating competing performance (with varying attributes) to the original multi-chart 3D GAN.

\subsection{Qualitative Results}

\begin{figure*}[t]
\includegraphics[width=\linewidth,trim={0 0 0 0}, clip]{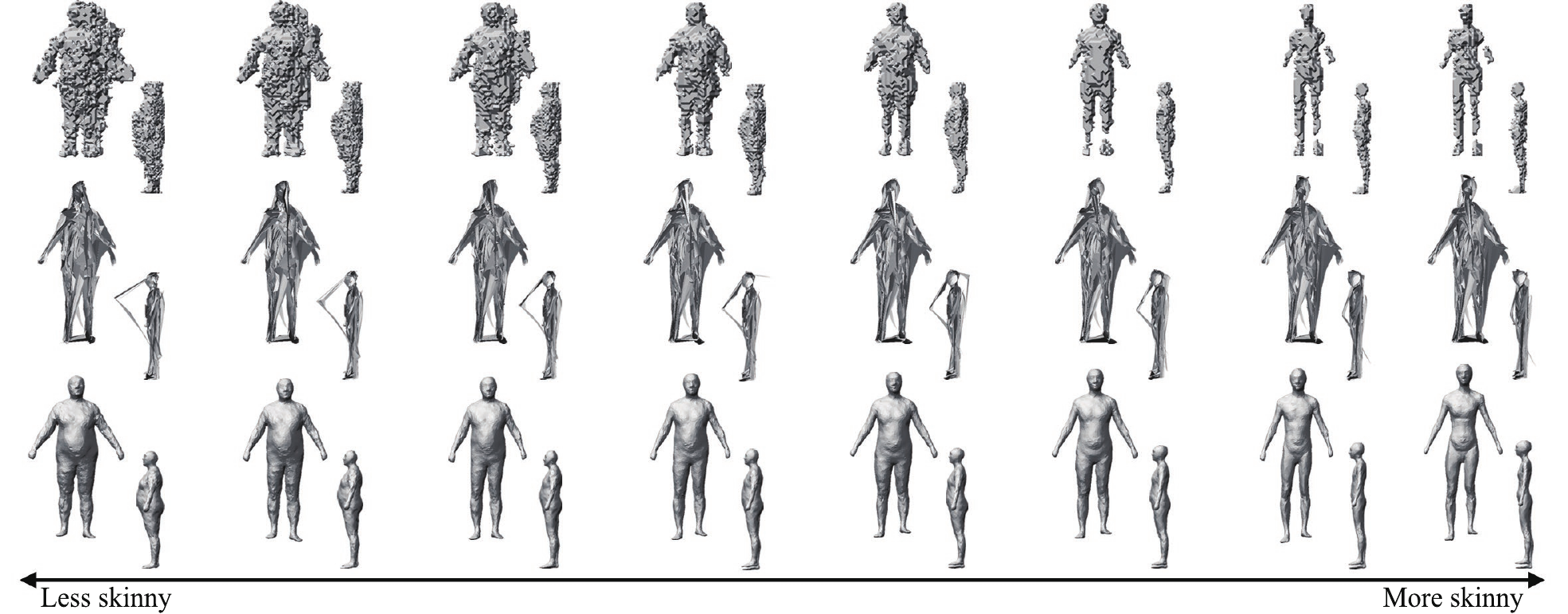}
\centering
\vnudge
\caption{Generated interpolation of human shape with respect to \emph{weight} attribute in the following methods: \textbf{Top:} Voxel 3D GAN. \textbf{Middle:} AtlasNet. \textbf{Bottom:} Our method Rank3DGAN.}
\label{fig:quality}
\vnudge
\end{figure*}

To compare our method with recent 3D generative model approaches, we extended these models to the conditional setting using semantic attributes, including the voxel 3D GAN \cite{3DGAN}, and AtlasNet \cite{AtlasNet} generative models. For the former, we have incorporated a ranking network and trained the whole architecture similarly to RankCGAN \cite{RankCGAN} procedure. For the latter, we focused on the task of mesh generation from point clouds input and concatenated the latent variable $r$ to the latent shape representation. We also added the ranking loss to the global objective function, while the ranker is a CNN that takes the point cloud coordinates and produces a ranking score.

We trained these models along with Rank3DGAN for $300$ epochs using $5000$ meshes, $5000$ pairwise comparison meshes of the human shape dataset \cite{HumanShape} and their corresponding point clouds with respect to \emph{weight} attribute. Figure \ref{fig:quality} shows the comparison between the selected methods.

Note that the voxel 3D GAN learned the desired transformation, but the result is of low resolution ($64^3$ dimensions) without surface connectivity and continuity compared to the mesh generated by our method. For AtlasNet interpolation, we remarked that the obesity of human mesh is represented by some vertices getting farther of the mesh. We alleged this behaviour to the fact that the objective function is optimal in this scenario where the resulting mesh is a close reconstruction to the input point cloud while the \emph{weight} ranking constraint is satisfied by moving some vertices outside the 3D shape. The interpolation failed to provide the desired output as we do not have access to the ground truth point cloud output given an input point cloud and a specific latent variable $r$ value.

\subsection{Applications}

\begin{figure*}[t]
\includegraphics[width=\linewidth,trim={0 0 0 0}, clip]{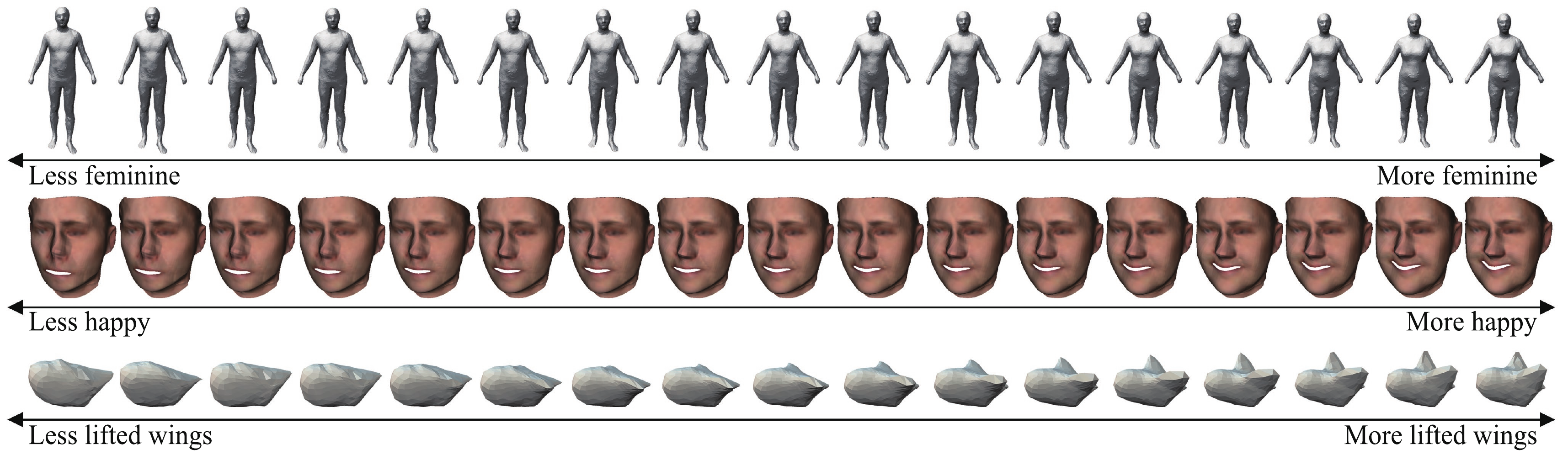}
\centering
\vnudge
\caption{Generated mesh interpolation in the following three datasets: \textbf{Top:} human shape, \textbf{Middle:} face, \textbf{Bottom:} bird with respect to the attributes \emph{weight, happy, lifted wings} respectively. }
\label{fig:generation}
\vnudge
\end{figure*}

\begin{figure*}[t]
\includegraphics[width=\linewidth,trim={0 0 0 0}, clip]{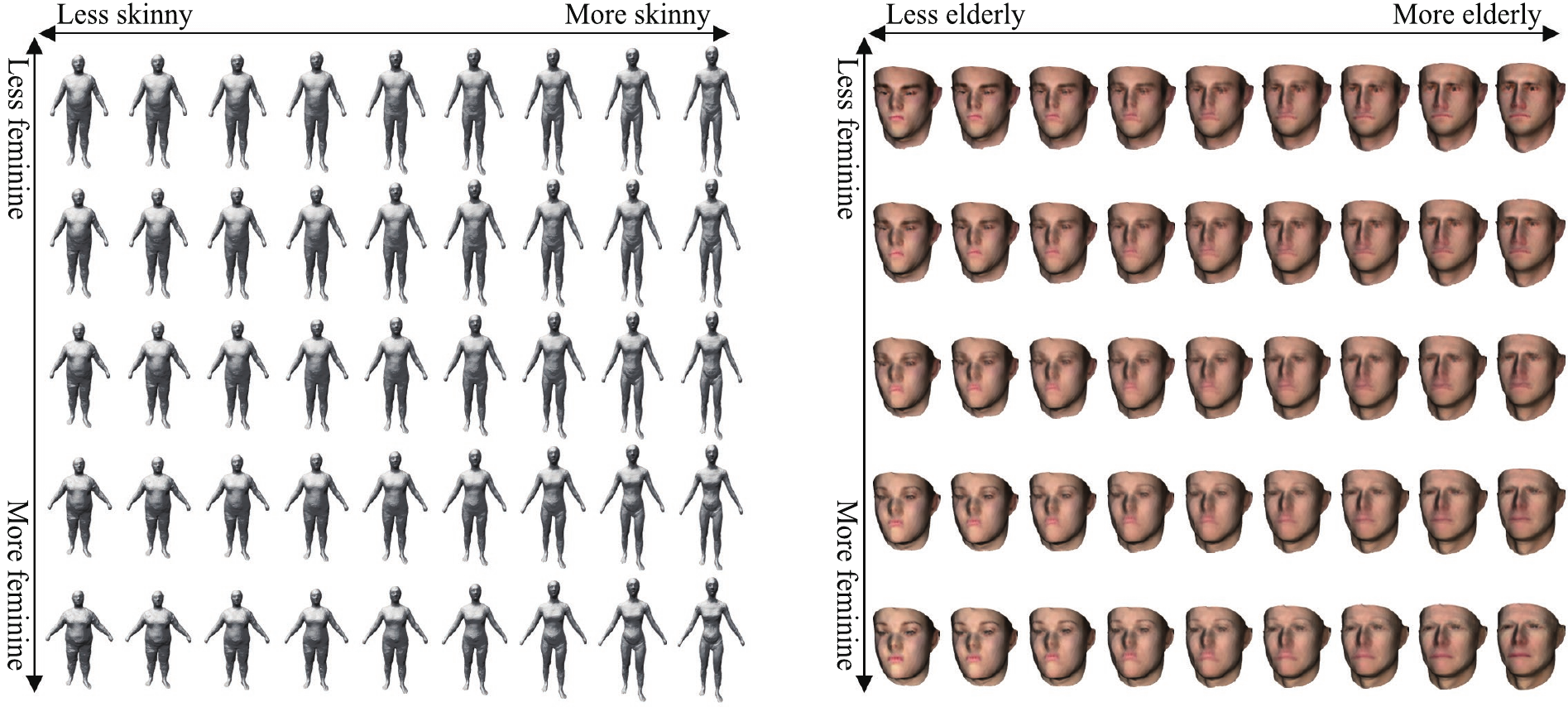}
\centering
\vnudge
\caption{Generated 2D mesh interpolation in the following two datasets: \textbf{Left:} human shape with respect to (\emph{weight, gender}) attributes and \textbf{Right:} face with respect to (\emph{age,gender}) attributes.}
\label{fig:2Dgeneration}
\end{figure*}

~~~~\textbf{Mesh generation:} To demonstrate the generative capability of our model, we choose single and double semantic attributes that span an interpolation line and plane respectively, where we fix the latent vector $z$ and vary the latent relative variables to each edge of the interval $[-1,1]$ in order to change the desired relative attributes. Figure \ref{fig:generation} shows how the generated meshes vary with respect to the value of the following semantic variables: \emph{weight, lifted wings, happy}. Figure \ref{fig:2Dgeneration} focuses on mesh interpolation with respect to two attributes. We select \emph{weight, gender} for human meshes and \emph{gender, age} for face meshes. We remark that the generated meshes are smooth and the attributes variation in the meshes is coherent with the selected relative attributes.



\begin{figure}[t]
    \begin{minipage}[t]{0.58\textwidth}
        \centering
        \raisebox{0.5\height}{\includegraphics[width=\linewidth]{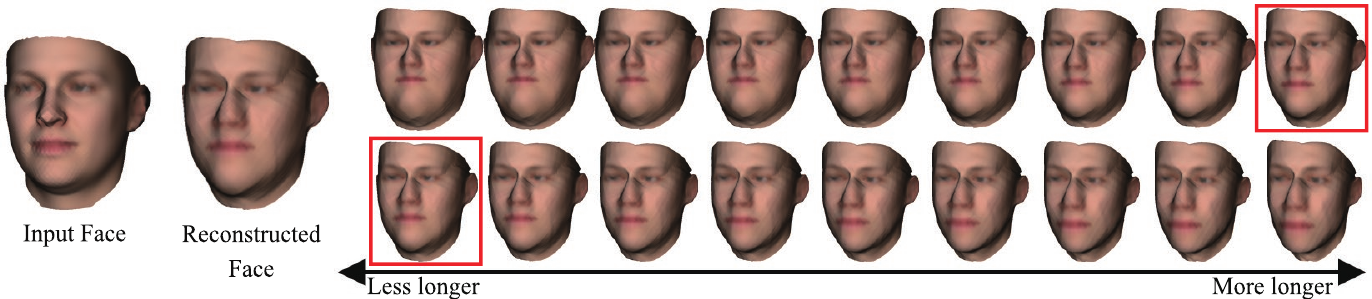}}
        \vnudge
        \caption{Mesh editing example in the face dataset with respect to \emph{height} attribute.}
        \label{fig:editing}
    \end{minipage}
    \hspace{0.01\linewidth}
    \begin{minipage}[t]{0.4\textwidth}
        \centering
        \includegraphics[width=\linewidth]{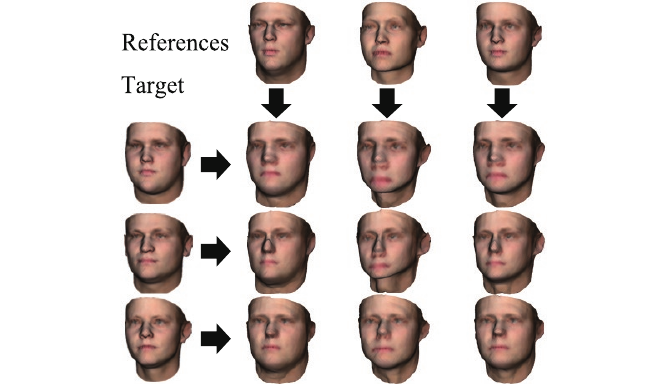}
        \vnudge
        \caption{Attribute transfer in the face dataset with respect to \emph{weight} attribute.}
        \label{fig:transfer}
    \end{minipage}
    \vnudge
\end{figure}

\textbf{Mesh editing:} This is an interesting feature for 3D object software tools. It enables the 3D graphic designer to alter high-level attributes in the mesh, that can be customised according to the user intent. This application consists of mapping an input mesh onto the subjective scale by estimating its reconstruction using the latent variables $r, z$, and then editing the reconstructed mesh by changing the value of $r \in [-1,1]$. Figure \ref{fig:editing} highlights this application in face meshes using \emph{height} attribute. The framed mesh is the reconstructed input mesh with latent variables $r^{*},z^{*}$, while $z^{*}$ is fixed, the first and second rows are the interpolation of $r$ in the range $[-1,r^{*}]$ and $[r^{*},1]$ respectively, denoting generated meshes of subjectively less or more emphasis on the chosen attribute.


\textbf{Mesh attribute transfer:} We can also transfer the subjective emphasis or strength of an attribute from a reference mesh to a target mesh. Using the encoder $E$, we quantify the semantic latent variable $r$ of the reference mesh, and then we edit the target mesh with the new estimated semantic value. Figure \ref{fig:transfer} shows this application for 3D faces in the case of \emph{weight} attribute. 

\section{Discussion and Conclusion}

We introduce Rank3DGAN, a GAN architecture that synthesises 3D meshes based on their multi-chart structure representation while manipulating local semantic attributes defined relatively using a pairwise comparisons of meshes. We have shown through experimental results that our model is capable of controlling a variety of semantic attributes in generated meshes using a subjective scale.
The main limitations of our model are inherited from multi-chart 3D GAN \cite{Multi-Chart}, which are the restriction to zero-genus surfaces, \emph{e.g.} sphere-like or disk-like surface and the usage of a fixed template mesh for reconstructing the mesh from the generated charts. Moreover, since the charts represent different parts of the mesh, the pre-processing chart normalization implicitly induces a loss of some global mesh attributes such as the \emph{height} of human shape, which complicates the task of learning a subjective measure using the ranking function. Possible extension of this work can focus on finding a novel way to represent global attributes so that it will not be lost in the mesh processing step. Another line of work can concentrate on enhancing the quality of the generated meshes by generating high resolution charts using other extensions of GANs or using advanced sampling and reconstruction techniques of the 2D flat-torus. Finally, other interesting applications can be derived from this work, such as 3D style transfer and 3D mesh reconstruction from RGB images.

\section{Acknowledgements}
We thank Wenbin Li, Christian Richardt, Neill D.F. Campbell for the discussion about the main ideas of the paper and Kwang In Kim, James Tompkin for the inspiration. Yassir Saquil thanks the European Union's Horizon 2020 research and innovation programme under the Marie Sk\l{}odowska-Curie grant agreement No 665992 and the UK's EPSRC Centre for Doctoral Training in Digital Entertainment (CDE), EP/L016540/1. 

\vfill
\bibliographystyle{ieee}
\bibliography{bib}

\clearpage
\appendix
\begin{appendices}
\section{Details on mesh parameterization for 2D CNN adaptation}
\label{sec:param}

\subsection{Seamless toric covers}
\begin{figure*}[h]
\includegraphics[width=0.95\linewidth]{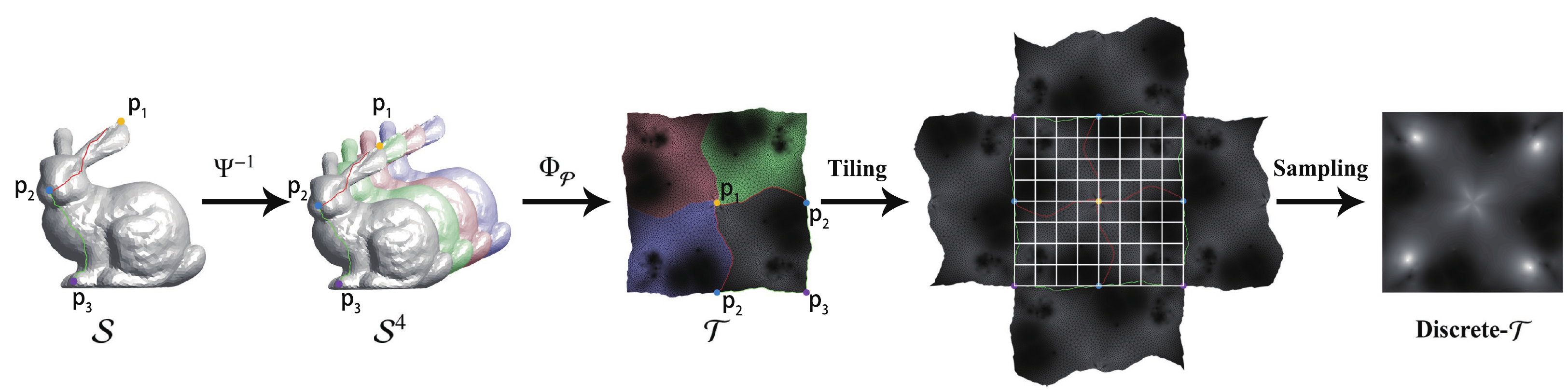}
\centering
\caption{The process of computing a discretized chart from a sphere-like 3D mesh. Where $\mathcal{S}$ the original surface, $\mathcal{S}^4$ the 4-cover of the surface, $\mathcal{T}$ the 2D flat-torus, and discrete-$\mathcal{T}$ the obtained chart after sampling and tiling the torus to fit a square. Since the parameterization is bijective, a reconstruction of the original mesh can be performed from the resulting chart.}
\label{fig:chart}
\end{figure*}

Formally, in order to build a seamless mapping between mesh surface $\mathcal{S}$ and flat-torus (toric) domain $\mathcal{T}$ , an intermediate surface, $\mathcal{S}^4$, of 4-cover of $\mathcal{S}$ is created as follows; four copies of the surface are cut along the path $p_1 \to p_2 \to p_3$ designed by triplet of points $\mathcal{P} = \{p_1,p_2,p_3\} \subset \mathcal{S}$ to get disk-like surface; Then, the four surfaces are stitched to get a surface torus, $\mathcal{S}^4$; Afterwards, the conformal mapping $\Phi_{\mathcal{P}} : \mathcal{S}^4 \to \mathcal{T}$ is computed using the parameterization method \cite{Tutte}; Lastly, the flat torus is tiled to cover the discretization grid resulting into a sampled chart as shown in the Figure \ref{fig:chart}.

\begin{figure*}[h]
\includegraphics[width=\linewidth]{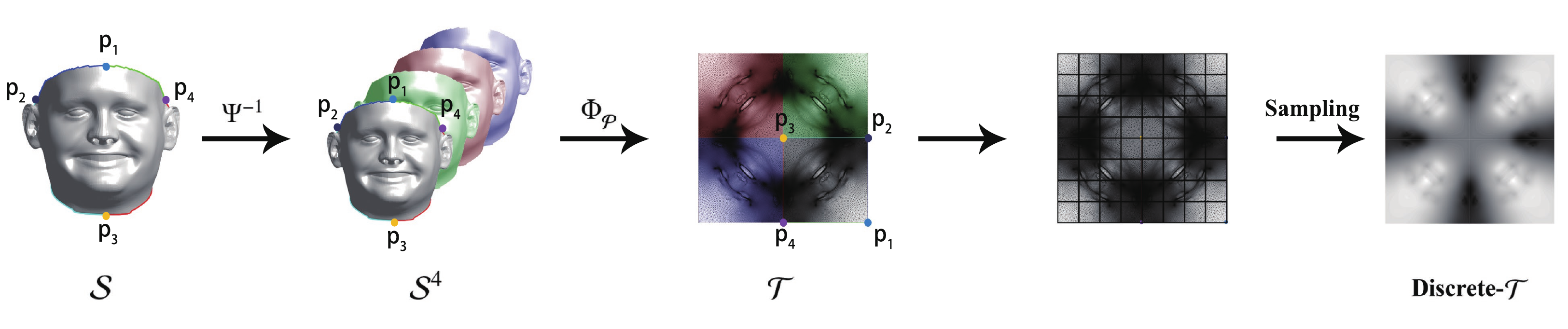}
\centering
\caption{The process of computing a discretized chart from a disk-like 3D mesh. Where $\mathcal{S}$ the original surface, $\mathcal{S}^4$ the 4-cover of the surface, $\mathcal{T}$ the 2D flat-torus, and discrete-$\mathcal{T}$ the obtained chart after sampling the torus. The tiling operation is not needed since the obtained torus is already a square.}
\label{fig:chart_face}
\vnudge
\end{figure*}

In the case of $\mathcal{S}$ being a disk-like surface, \emph{e.g.} 3D face meshes, a surface cut is not needed and a quadriplet of points $\mathcal{P} = \{p_1, p_2, p_3, p_4\} \in \partial\mathcal{S}$ is designated, with $\partial\mathcal{S}$ the surface boundary, in order to stitch the four surfaces to get a surface torus as shown in the Figure \ref{fig:chart_face}.

$\text{push}_{\mathcal{P}}(x) = x \circ \Psi \circ \Phi_{\mathcal{P}}^{-1} \in \R^{c\times n \times n}$ is defined as the obtained conformal chart after transferring a function $x \in \mathcal{F}(\mathcal{S},\R^c)$ over the surface to a flat-torus using $\Phi_{\mathcal{P}}$, where $\Psi : \mathcal{S}^4 \to \mathcal{S}$ is the projection of 4-cover surface to the original surface. A conformal chart produces an area scaling with respect to the selected triplet $\mathcal{P}$, thus a careful choice of multiple triplets can ensure a global coverage of the surface with a minimum distortion.

\subsection{Multi-Chart Structure}

The multi-chart structure \cite{Multi-Chart} is a set of charts that globally cover the surface of mesh. Formally, it is a tuple $(P, F)$, where $P \in \R^{n \times 3}$ is the set of landmarks and $F$ is the set of landmark triplets, such that each triplet $\mathcal{P} = \{p_i,p_j,p_k\} \in F$ represents a chart and every mesh $S_i$ consists of $|F|$ charts.

Along with the covering property, the multi-chart structure should ensure the scale-translation (s-t) rigidity \cite{Multi-Chart} that allows unique recovering the original scale and mean of the charts after the normalization applied while training a network.

Recall that our purpose is to learn a generative model for 3D meshes. Which leads us to consider the coordinates $X = (x, y, z)$ as functions over the mesh $X \in \mathcal{F}(\mathcal{S},\R^3)$ to be transferred using the multi-chart structure $(P,F)$ as follows:
\begin{equation}
    \text{push}_{\mathcal{P}}(X) = X \circ \Psi \circ \Phi_{\mathcal{P}}^{-1} \in \R^{3 \times n \times n}
\end{equation}
Where $\text{push}_{\mathcal{P}}(X)$ is the obtained chart with spatial dimension $n \times n$ using the triplet $\mathcal{P} \in F$. By concatenating the charts obtained from using all triplet in $F$, we get the final multi-chart tensor representing the whole mesh defined by: $C \in \R^{3|F| \times n \times n}$. Since every $3$ charts in $C$ represent a different part in the mesh, we need to normalize them for an optimal learning process. In the end, the scale-translation rigidity property ensures a unique reconstruction.

Thanks to the multi-chart structure, learning a generative model $G : \R^d \to \R^{3|F| \times n \times n}$ for 3D meshes can be performed in the image space using GANs \cite{GAN} with few considerations for the geometric setting described below as geometric consistency layers:

\begin{itemize}
    \item Standard convolution and deconvolution are substituted by cyclic-padding ones as suggested in \cite{Toric,Multi-Chart} to take into account the invariance to torus symmetry.
    \item Projection layer is incorporated in the generator to satisfy the invariance of $\mathcal{S}^4$ symmetries.
    \item Landmark consistency layer \cite{Multi-Chart} is incorporated to enforce the generated normalized charts $G(z) \in \R^{3|F| \times n \times n}, z \in \R^d$ to satisfy the s-t rigidity property.
    \item Zero-mean layer is proposed to reduce the mean of every generated chart after applying the landmark consistency layer.
\end{itemize}

The final step following learning a generative model $G$ is to reconstruct the 3D mesh from the generated charts $C = G(z) \in \R^{3|F| \times n \times n}$ using template fitting \cite{Multi-Chart}.

Since $C$ is output of the landmark consistency layer, we can uniquely recover the scale and mean of charts resulting in charts denoted by $\hat{C}$. Afterwards, a template mesh $\mathcal{S}^t$ is used to obtain the generated mesh by setting the mesh vertices location using the multi-chart structure $(P,F)$ and the generated charts $\hat{C}$ as follows:

\begin{equation}
    v = \frac{\sum_{\mathcal{P} \in F} \tau_{\mathcal{P}}(v)~ \hat{C}(\Phi_{\mathcal{P}}(v))}{\sum_{\mathcal{P} \in F} \tau_{\mathcal{P}}(v)}
\end{equation}
\\
Where $\tau_{\mathcal{P}}(v)$ is the inverse area scale of the 1-ring of vertex $v$ exhibited by flat-torus $\Phi_{\mathcal{P}}(\mathcal{S}^t)$ of the template mesh and $\hat{C}(\Phi_{\mathcal{P}}(v))$ is the image of the flat-torus vertex $v$ under the learned charts in $\hat{C}$ associated with triplet $\mathcal{P}$, which is computed using bilinear interpolation of the grid cells in these learned charts.

\section{Details on data preparation}
\label{sec:appendix-data}

MPII Human Shape \cite{HumanShape} is a statistical body representation learned from CAESAR dataset \cite{CAESAR}, which describes body variations in a set of spaces containing 20-dimensional linear shape basis. In order to obtain human body meshes dataset, we sampled $15000$ shapes from the first $10$ PCA components of the set of spaces, where each component is scaled by a random value sampled from $\mathcal{U}(-3\sigma,3\sigma)$, with $\sigma$ the standard deviation of a specific component. Afterwards, we built $5000$ pairwise comparisons as follows: We first manually selected the PCA component that exhibits a strong variation in the desired attribute and we sampled uniformly random scales from $\mathcal{U}(-3\sigma,3\sigma)$ for the remaining components. Then, we sampled a pair of meshes by sampling the corresponding scale of the selected component from $\mathcal{U}(-3\sigma,-2\sigma)$ and $\mathcal{U}(2\sigma,3\sigma)$ respectively. 
Finally, we obtained the relative attributes \emph{weight, gender} by selecting the second, third or/and fourth PCA components respectively. 
Concerning the case of multiple attributes, we require multiple attributes for a specific pair of meshes. For this reason, we built $5000$ pairwise comparisons similarly as described previously, but instead of selecting only one component that represents the desired attribute, we selected multiple components that represent all desired attributes in the same time. In order to construct the multi-chart structure, we selected manually $21$ landmarks and $16$ triplets for better body shape coverage.

CUB-200-2011 dataset \cite{CUB_200_2011} contains $11788$ images of 200 species of birds. We followed \cite{Bird} in order to reconstruct 3D meshes from CUB-200-2011 dataset \cite{CUB_200_2011}. We then randomly selected $5000$ meshes after smoothening the obtained meshes and filtering out the ones with intersected faces for a better mesh quality. In addition, we built $5000$ pairwise mesh comparisons with respect to \emph{lifted wings} attribute, where this attribute is defined by comparing the angle made by the wings extremum vertices and the neck vertex between a pair of meshes. Finally, for the multi-chart construction from bird meshes, we selected manually $14$ landmarks and $11$ triplets to cover the whole bird mesh.

Basel Face Model is a 3D morphable face model with two versions. Basel Face Model 2009 \cite{Basel09} provides different face identities by combining 199 PCA components. Since the model was built on top of labelled training data, it provides face coefficients along the directions of maximal variance for an attribute. We relied on PCA components to sample $5000$ face meshes, then we built $5000$ pairwise mesh comparisons by sampling along the directions of maximal variance for each of the following attributes: \emph{gender, height, weight,} and \emph{age}.
Similarly, Basel Face Model 2017 \cite{Basel17} provides different face identities and expressions by combining 199 and 100 PCA components respectively, where we also sampled $5000$ face meshes and $5000$ pairwise comparisons for each of the following provided attributes: \emph{anger, disgust, fear, happy, sad, surprise}.
In the multiple attributes case, we build $5000$ pairwise comparisons by sampling simultaneously along the corresponding PCA components of the desired attributes.
We preprocessed these datasets to ensure that the meshes are zero-genus by stitching the mouth using additional faces and keeping consistent number of vertices. When we reconstruct the mesh, we delete the added faces as a postprocessing step to recover the original mesh.
Finally, since the 3D faces are disk-type surfaces, a single chart is enough to cover the whole surface by manually selecting $4$ landmarks and $1$ quadriplet.
\end{appendices}

\end{document}